\documentclass[runningheads]{llncs}
\usepackage{wrapfig}
\usepackage{graphicx}
\usepackage[caption=false]{subfig}
\begin{document}
\title{Activity Recognition and Prediction in Real Homes\thanks{Financed by the Norwegian Research Council under the SAMANSVAR programme (247620/O70).}}
\titlerunning{Activity Recognition and Prediction in Real Homes}
%

\author{Fl\'{a}via Dias Casagrande\and Evi Zouganeli}
\authorrunning{F.D. Casagrande and E. Zouganeli.}
%
\institute{OsloMet -- Oslo Metropolitan University, Pilestredet 35, 0166 Oslo, Norway \\
\email{\{flacas,evizou\}@oslomet.no}}
\maketitle              
\begin{abstract}
In this paper, we present work in progress on activity recognition and prediction in real homes using either binary sensor data or depth video data. We present our field trial and set-up for collecting and storing the data, our methods, and our current results. We compare the accuracy of predicting the next binary sensor event using probabilistic methods and Long Short-Term Memory (LSTM) networks, include the time information to improve prediction accuracy, as well as predict both the next sensor event and its time of occurrence using one LSTM model. We investigate transfer learning between apartments and show that it is possible to pre-train the model with data from other apartments and achieve good accuracy in a new apartment straight away. In addition, we present preliminary results from activity recognition using low resolution depth video data from seven apartments, and classify four activities -- no movement, standing up, sitting down, and TV interaction -- by using a relatively simple processing method where we apply an Infinite Impulse Response (IIR) filter to extract movements from the frames prior to feeding them to a convolutional LSTM network for the classification. 
\end{abstract}
\section{Introduction}

The Assisted Living project is an interdisciplinary project with expertise in the fields of smart-home technology, machine learning, nursing and occupational therapy, and ethics. The aim is to develop assisted living technology (ALT) to support older adults with mild cognitive impairment or dementia (MCI/D) live a safe and independent life at home \cite{Zouganeli2017}. MCI and dementia involve a cognitive decline that can affect attention, concentration, memory, comprehension, reasoning, and problem solving. A number of research studies have investigated functions in smart-home environments to support older adults in general, and those with MCI/D in particular, in their everyday life. These include assisting functions such as prompting with reminders or encouragement, diagnosis tools, as well as alarm creation, prediction, anticipation, and prevention of hazardous situations. The majority of these functions requires reliable activity/ action recognition and prediction algorithms to work properly. This field is at a quite early stage at the moment. With the exception of fall detection, there are currently no commercial systems with such functionality nor are there any complete prototypes available at research and development level. 

The aim of our work is to use activity prediction to realize support functions for older adults with MCI/D. In this paper we present work in progress on action/ activity recognition and prediction using data from real homes, seven apartments, each with one older adult resident over 65 years old –- the majority over 80 years old. We use binary sensors as well as a low resolution depth video camera that is in fact a commercial fall detection system called RoomMate \cite{rm}. We present results on activity prediction based on the binary sensors, where we compare probabilistic methods and neural networks, include the time information and predict the time of occurrence as well as the next sensor event, and investigate transfer learning between apartments. In addition, we present preliminary results from action recognition based on video frames that contain movement information. 

\section{Related Work}

\subsection{Activity Prediction using Binary Sensors}

Several sequential data prediction algorithms have been investigated in the past years \cite{Wu2017}. The Active LeZi (ALZ) is a probabilistic method that has been extensively employed for prediction on sequential data \cite{Cookc}. Based on the ALZ, the Sequence Prediction via Enhanced Episode Discovery (SPEED) algorithm was implemented \cite{Alam2012a}. Recurrent neural network (RNN) models -- Echo State Network (ESN), Back Propagation Through Time (BPTT), and Real Time Recurrent Learning (RTRL) -- were applied and the ESN performed better when predicting the next sensor in a sequence \cite{Lotfi2012}. 

Activity prediction includes mainly two tasks: sequence prediction and time prediction. In addition to sequence prediction, the algorithms mentioned above should also be able to predict when the next symbol (representing either a sensor or an activity) will occur.  Several algorithms have been used to predict the time of occurrence alone, such as the time series methods Autoregressive Moving Average (ARMA) and Autoregressive Integrated Moving Average (ARIMA) \cite{pelham1990}; non-linear Autoregressive Network (NARX), Elman network to predict a sensor activation’s start and end time \cite{Mahmoud2013}; decision trees \cite{Minor2016}; Poisson process \cite{poisson2016}. 

To our knowledge, only one work predicts both sensor event and time in the same model \cite{He2015}. This work uses a Bayesian network and predicts the next location, time of day (slots of 3 hours through the day), and day of the week with reported accuracies of 46-60\%, 66-87\% and 89-97\%. Subsequently, the activity is predicted with an accuracy of 61-64\% based on a combination of these features. They use data from testbeds. Our dataset was collected from a real home, it contains events from fifteen binary sensors, i.e. twice as many as used in \cite{Lotfi2012,Mahmoud2013}, less than one third of the number of sensors used in the Mavlab testbed \cite{Minor2016,Cookc,Alam2012a}, and half of \cite{He2015}. Our work predicts the next sensor event and the time of occurrence for a set with 15 sensors in the same model using LSTM networks.

\subsection{Activity Recognition using Depth Video Data}

There is strong evidence that technology can support aging at home \cite{reeder} and a large number of studies have implemented assistive technology to support older adults live a safe and independent life at home \cite{Blackman2016}. Human activity recognition (HAR) has been well studied in the past years \cite{Zhang2017} and a number of algorithms have been used. Hidden Markov Models (HMM) achieved a maximum accuracy of 97\% with the MSR Action3D dataset with skeleton data histograms fed to a HMM \cite{Xia2012a}. Yang et al.~\cite{Yang2012b} reached 97\% accuracy on the same dataset by using Depth Motion Maps and Histogram of Oriented Gradients (DMM-HOG) features and SVM. Convolutional neural networks have achieved remarkable results for HAR from depth data. Wang et al.~\cite{wangaction} achieved 100\% accuracy on the same action dataset with a deep convolutional network by using weighted hierarchical DMMs of the video sequences.

\section{Field Trial}

\begin{figure}[h]
\centerline{\includegraphics[width=0.4 \textwidth]{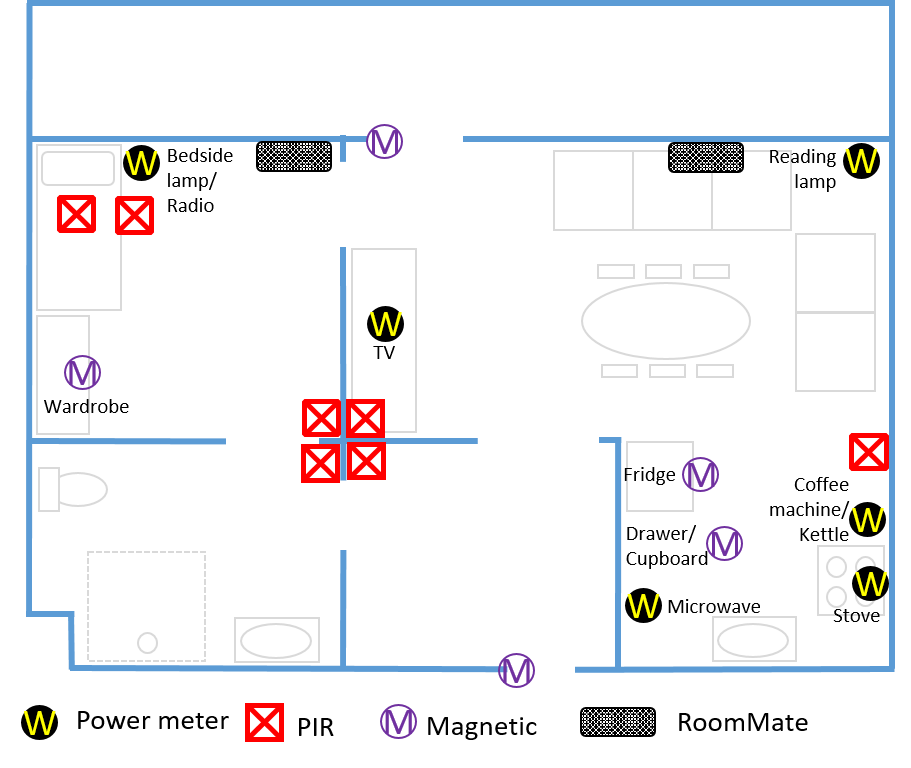}}
\caption{Sensors system installed in the field trial apartment.}
\label{apt}
\end{figure}

Our field trial involves seven independent one-bedroom apartments within a community care facility for people over 65 years old. Each apartment comprises a bedroom, a living room, open kitchen area, a bathroom, and an entrance hall (Fig. \ref{apt}). Our set of sensors contains motion, magnetic, and power sensors. These enable inference of occupancy patterns (movement around the apartment) and some daily and leisure activities. Unfortunately, not all apartments could have the exact same set of sensors due to physical limitations (e.g. fridge door with a too big gap to enable the use of magnetic sensor) and/ or different equipment (e.g. residents either have a coffee machine or a kettle). However, all the participants had the same initial proposal of set of sensors, as shown in Fig.~\ref{apt}. There are two RoomMate depth video cameras in each apartment, one in the living room and kitchen area, the other in the bedroom area, as shown in Fig. \ref{apt}. The RoomMate is an infra-red (IR)-based depth sensor and measures the distance of surfaces to the camera by time-of-flight (TOF) technology. The resolution is 160x120 pixels, with a rate of 25 frames per second. This is rather low resolution -- a fact that is advantageous with respect to privacy, but makes data processing quite challenging.

\begin{table}[h]
  \caption{Binary sensors data}\label{tab:data}
  \centering
  \begin{tabular}{| c | c | c |} 
  \hline
      \emph{Timestamp} & \emph{Sensor ID} & \emph{Sensor message} \\
      \hline
      01.09.2017 07:58:40 & 4 & 1\\
      01.09.2017 07:59:02 & 10 & 1 \\
      01.09.2017 08:03:05 & 10 & 0\\ 
      \hline
  \end{tabular}
\end{table}

\section{Activity Prediction using Binary Sensors}

\subsection{Data Preprocessing}

The preparation of the binary data includes two steps: data correction and data conversion. The data correction is necessary because the data acquired from binary sensors often contain faulty events \cite{Elhady}. In our system, occasionally the motion sensors do not send an activation event when they should. Missing sensor events have been inserted to correct for this. For example, it is not possible to go to the bedroom directly from the kitchen without passing through the living room. When the living room activation event is missing, it is inserted. The time of the inserted event is the mean between the previous and next event. This does not compromise the dataset accuracy because the faulty events are usually between relatively fast motions around the apartment, hence the elapsed time is short. Subsequently, the corrected data is converted to sequences of letters. This is inspired by the SPEED algorithm \cite{Alam2012a}. Upper- and lower- case letters represent a sensor's ``on" and ``off" events. For the sample data in Table~\ref{tab:data}, SPEED would generate the sequence ``ABb", where sensors 4 and 10 are assigned the letters a/A and b/B, respectively. Afterwards, we include the time information. This is done in two ways as follows. In all cases the generated sensor events are treated as independent events, as presented also in the next section.

\subsubsection{Sensor Event with Elapsed Time Classes.} Here we use two fixed sets of time intervals: [$<$1min, 1-15min, 15min-1h, $>$1h] and [$<$1min, 1-5min, 5-15min, 15-30min, 30min-1h, 1-2h, 2-5h]. This results in a 4-class case and an 8-class case.

\subsubsection{Sensor and Time-Cluster with Hour of the Day and Elapsed Time to the Next Event.} We apply the K-means algorithm to cluster each sensor event according to the hour of the day occurrence and the time elapsed to the following sensor event. In the K-means algorithm the samples of each sensor are classified into K clusters such that the sum of square distances (SSD) within the clusters is minimized \cite{Bataineh2011ACS}. Each cluster contains a centroid, given by the mean value of each feature of the algorithm.  We perform K-means for a maximum number of clusters (K) equal to 8 and choose the best K manually according to the elbow method \cite{Joshi2013ModifiedKF}.

\subsection{Probabilistic Methods}

Both ALZ and SPEED translate the data acquired from the sensors into a sequence of letters and identify patterns that occur frequently \cite{Casagarnde,Casagrande2019}. The patterns and their frequency of occurrence are used to generate a tree, which is then used to calculate the next most probable event to occur based on the Prediction Partial Matching algorithm (PPM) \cite{Alam2012a}. 

\subsection{Long Short-Term Memory Network}

\begin{figure}[h]
\centerline{\includegraphics[width=1 \textwidth]{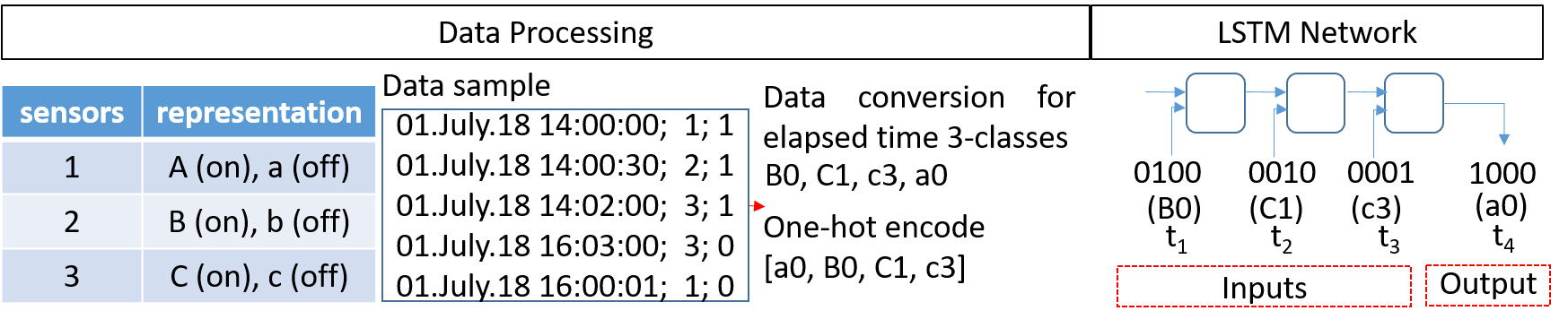}}
\caption{Configuration of LSTM network.}
\label{config}
\end{figure}

RNN has been broadly applied to sequence prediction due to its property of keeping an internal memory. Hence, it attains a good performance for inputs that are sequential in time. The LSTM is an RNN architecture designed to be better at storing and accessing information than the standard RNN \cite{Hochreiter1997}. In this work the LSTM network is configured as a text generation network. The number of inputs is a certain number of symbols (sensor events with numbers indicating time) –- equal to the memory length -– and the output is the predicted next symbol in the sequence (Fig.~\ref{config}). The input and output are one-hot encoded. Hence, our input vector has as many values as the number of symbols in the sequence. In the case of 15 sensors, we have 30 inputs to represent the ``on" and ``off" states of each of these. E.g. when the 4-class interval is taken into account, the number of inputs is multiplied by 4 (120 inputs in total) and similarly in the other cases.

A stateless LSTM network model was implemented in Python 3 using Keras open source library for neural networks. The memory length (i.e. number of previous events used to predict the next event) \cite{FlaviaDiasCasagrandeJimTørresen2018} had value 10. The model has one hidden layer with hyperbolic tangent activation and 64 neurons. Our batch size (i.e. number of samples used for training each iteration of the epoch) was 512. We used Adam as the optimization function with learning rate of 0.01 and categorical cross-entropy as loss function. The output layer was a softmax activation function. We used the early stopping method to avoid overfitting and unnecessary computations, allowing a maximum of 200 epochs for each model's training.

\subsection{Sensor Event Prediction using Binary Data}

\begin{figure}[h]
\centerline{\includegraphics[width=0.5 \textwidth]{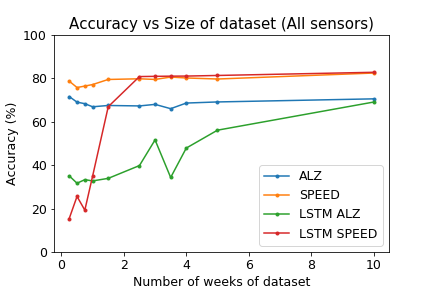}}
\caption{Accuracy vs. size of dataset for all algorithms.}
\label{sizeAll}
\end{figure}

In this section we compare four methods, two probabilistic and two neural networks. In all cases, the results show the mean accuracy achieved using a 5-fold cross-validation process (using 60\% of the data for training, 20\% for validation, and 20\% for testing). We investigate the dependence of the accuracy on the size of the dataset used for the complete process of training, validating, and testing the models. Prior to this, the optimum number of previous events to base the prediction of the next event on is found for each of the methods \cite{Casagrande2019}. The accuracy results are computed within the testing set using the optimal memory length for each method. Fig.~\ref{sizeAll} shows the results when the algorithms are applied to a dataset from a single apartment. A peak accuracy of 83\% was achieved by LSTM with SPEED-text, while the SPEED algorithm achieved a peak accuracy of 82\%. The accuracy achieved by the LSTM with ALZ-text was considerably lower at 69\%. In this case, stability is achieved much later than with the other methods. Finally, the ALZ method reached a top accuracy of 70\% with 4 weeks of data. Note that the probabilistic methods attain a good accuracy (close to the peak accuracy) with only 2 days of data. By comparison, the LSTM networks need approximately 2-3 weeks of data to start approaching their top accuracy. This indicates that the LSTM can learn longer term patterns and dependencies, and attain better accuracy based on these. In addition, LSTM networks needed much shorter time to train than SPEED – eight times faster.

Subsequently, we develop the model further by using the time information in the best method –- LSTM with SPEED-text. Here, the LSTM network was trained based on a 3-fold cross-validation. We  use  in total 40 weeks with recorded data from one apartment where we apply our algorithms, which accounts for 163347 events. Some accuracy curves do not show significant improvement after a certain number of events, and we therefore show the plots up to a certain point for better clarity on the lower range of the graph.

\begin{figure}[h!]
\centerline{\includegraphics[width=0.7 \textwidth]{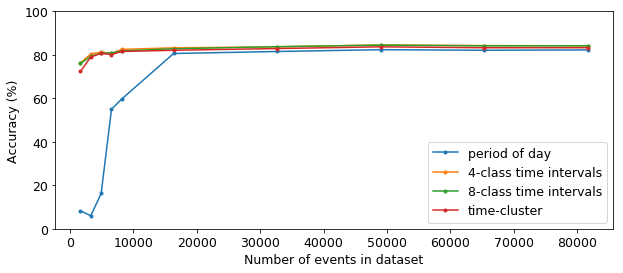}}
\caption{Accuracy of prediction of next sensor event vs. the number of events in the dataset.}
\label{predSensor}
\end{figure}

Firstly, we predict the next sensor event based on the two proposed input sequences with the time information (section 4.2). Fig.~\ref{predSensor} shows the performance of the prediction vs. the amount of data in the dataset. When we include the time information in the input, the accuracy is improved by 1-1.4\% for all methods as compared to Fig.~\ref{sizeAll}. The highest accuracy (84.39\%) is achieved by the 4-class time-interval. The small improvement compared to the best results from Fig.~\ref{sizeAll}, was initially somewhat surprising, however, it can be explained by the fact that the apartments are quite small and there is a limited number of sensors and alternative sequences. The choice of 4- or 8-class time-interval classes does not have a significant effect on the accuracy. This is presumably because most of the events have a short elapsed time to the next event.

\subsection{Prediction of Next Sensor Event and Time of Occurrence}

\begin{figure}[h]
\centerline{\includegraphics[width=0.7 \textwidth]{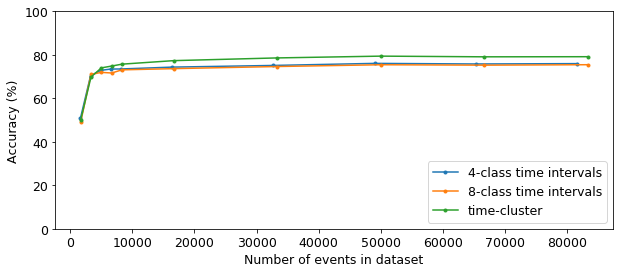}}
\caption{Accuracy of prediction of next sensor event and time information vs. the number of events in the dataset.}
\label{predSensorTime}
\end{figure}

We examine the accuracy of predicting both the next sensor event and time information using 4- and 8- class time-intervals or the K-means time-cluster. Lower accuracies are attained than when predicting only the next sensor event, as expected, since now more information is being predicted with the same model. The best accuracy was achieved by the K-means time-cluster (79.68\%), 4\% better than the second best-performing method (Fig.~\ref{predSensorTime}). The required number of events in the dataset is similar for the three methods, about 10000 events .

\subsection{Transfer Learning between Apartments}

\begin{figure*}[h]
\centering
   \subfloat[]{\label{graph4}
   \centering
      \includegraphics[width=.4\textwidth]{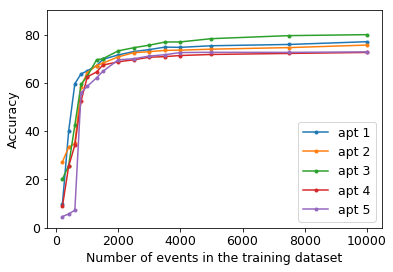}}
~
   \subfloat[]{\label{graph5}
\centering
      \includegraphics[width=.4\textwidth]{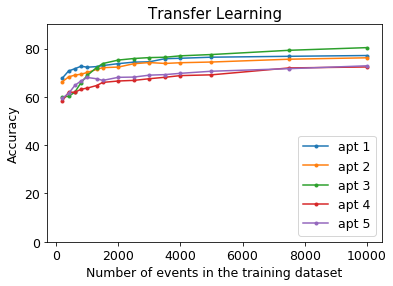}}

   \caption{Accuracy of prediction of the next sensor and time-cluster vs. number of events in the training dataset, using as input both sensor event and time-cluster, a) separately for each apartment, b) transfer learning, training the model with data from four apartments, fine-tuning with and testing on the target apartment.}\label{grapht}
\end{figure*}

As described in Section 3, some power and magnetic sensors differ within the five apartments. In order to transfer the learning across the apartments, we re-label the sensors that refer to the same activity. Lamp power sensor events were removed from the datasets since we did not manage to assign them to an activity that was common for all lamps and apartments.
In all cases, the LSTM network was trained based on a certain number of events and tested on a test set containing 3000 events. This process is repeated three times and the accuracy values in the graph correspond to the mean of the best test accuracy of each training. Fig.~\ref{graph5} presents the accuracy results when transfer learning is carried out as follows. The model is trained using data from four apartments and fine-tuned with and tested on the target apartment. A very low number of events is required for the fine-tuning to achieve quite high accuracy straight away. The accuracy increases slowly as more events are added for the fine-tuning.

\section{Activity Recognition using Depth Video Data}

\subsection{Data preprocessing}

Median filtering is applied to the raw depth video data to remove noise. The process consists of removing very low and very high pixel values in the image and replacing them with the median value of the nearest neighbors. A 5x5 filter was applied to each frame, as a compromise between image sharpness (quality) and its high frequency background noise. After this first step, we apply an Infinite Impulse Response (IIR) filter. The filter is configured as a first-order high pass in this work, which leads to capturing any movement. Finally, a last processing step is performed in order to normalize the length $n_i$ of frames of the video samples to a fixed length $N,$ as this is a prerequisite for the convLSTM model. Frames are deleted if the sequence is shorter than N, or inserted if the sequence is longer than $N.$ In both cases this takes place in equally spaced positions in the sequence (in accordance with the number of frames that need to be deleted/ inserted), $n_i=N.$ The value of each pixel in the inserted frames is equal to the mean between the preceding and succeeding frames. We use a convolutional long short-term memory network (convLSTM) for the classification. Convolutional neural networks (CNNs) have been widely used to process multiple arrays of data, including color or depth images. By combining CNNs with RNNs, the model is able to learn both spatial and temporal features from a sequence of frames \cite{Shi2015}.

In this work, the preprocessed and IIR-filtered frame sequences were labelled into four movement categories: TV-related movements (turn it on/ off and switch over channels), standing up, sitting down and no movement. After the preprocessing, the sequences were fed to a convLSTM. The first two steps were implemented by using the median and IIR filters from the SciPy library. The convLSTM was implemented using the Keras library. The trained model comprises one convLSTM layer with three 3x3 filters and hyperbolic tangent activation, followed by a dense layer with softmax activation. The batch size was 16 and learning rate 0.01. Optimization function Adam and loss function categorical cross-entropy were optimal. We set the dropout ratio to 0.5 in order to avoid overfitting, as well as early stopping.


\subsection{Activity Recognition}

A total of 800 video sequences (200 of each category) were extracted from recordings acquired from real homes, from seven different residents. Each video sequence is length normalized to a size of 100 frames. We split our dataset into training (80\%) and testing (20\%) sets. They are both balanced for all classes (i.e. equal percentage of samples per class in each set).

\begin{figure}[h]
\begin{center}
   \includegraphics[width=0.5\linewidth]{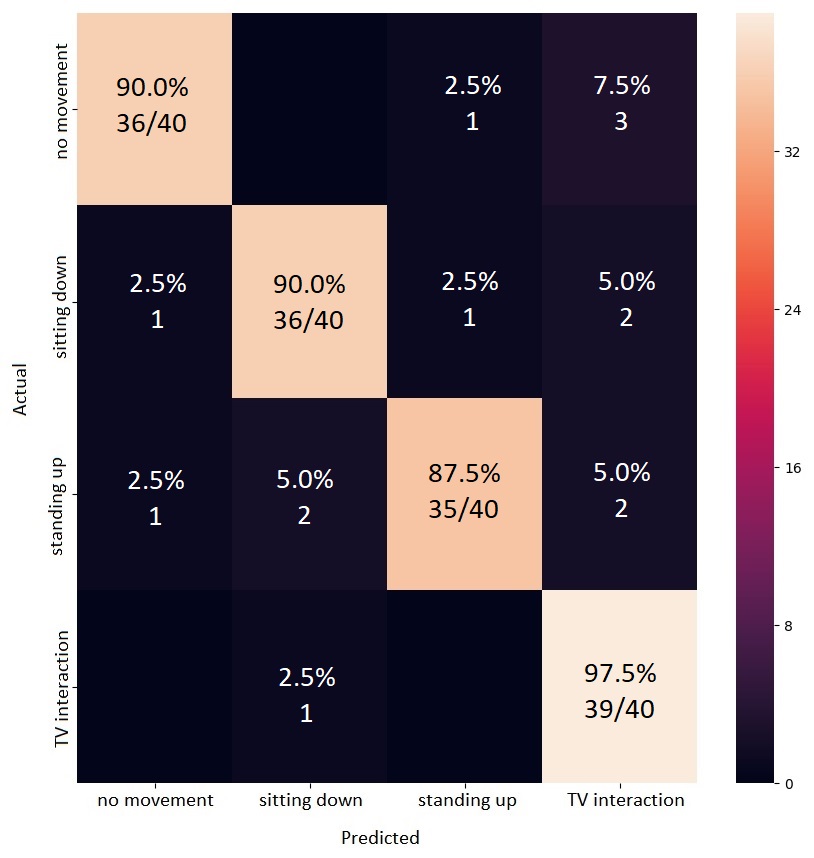}
\end{center}
   \caption{Confusion matrix of model for 800 samples using both median and IIR filter.}
\label{fig:200}
\end{figure}

We analyze the test accuracy attained for different sizes of datasets for two cases: only median filter, and both median and IIR filter. The obtained results correspond to an average of the three best accuracies achieved by different trained models -– shuffling the training and testing data. The use of the IIR filter resulted in a best average peak accuracy of 86.04\%, whereas by comparison without the IIR filter the best average peak accuracy achieved was 82.50\%. Using the IIR filter improves the accuracy by approximately 4\%, for all data sizes. The accuracy improves slowly as more samples are added, in both cases. The model did not reach stability with the available data, as the accuracy keeps increasing with data size. Hence better accuracy should be possible to achieve with additional samples. The confusion matrix is shown in Fig. ~\ref{fig:200}.


\section{Conclusion and Future Work}

In this paper, we present work in progress, our field-trial and set-up for collecting and storing the data from real homes, and our results on activity recognition and prediction using either binary sensor data or depth video data. We compare the accuracy of predicting the next binary sensor event using probabilistic methods and LSTM networks. LSTM achieved the best accuracy of 83\%. Using time information referring to the next sensor event improved this accuracy by 1.4\%. Finally, we predicted both the next sensor event and its time of occurrence with a peak average accuracy of 80\%. We have investigated transfer learning between apartments and shown that it is possible to pre-train the model with data from other apartments and achieve good accuracy straight away (70-80\%) from the first day. The top accuracy in this case is similar to the one achieved when training each apartment individually.

We have in addition shown preliminary results from activity recognition using low resolution depth video data from seven apartments. 800 video samples were extracted containing four classes: no movement, standing up, sitting down, and TV interaction. We use a relatively simple processing method where we apply an IIR filter to extract movements from the frames prior to feeding them to a convLSTM network for the classification. We achieved an overall mean peak accuracy of 86\%, with the accuracy of all classes reaching at least 85\%. The method managed to identify TV-interaction actions with a peak accuracy of 97.5\%. When the IIR filter is not used the accuracy is about 4-5\% lower.

Future work will use state-of-the-art video processing techniques to carry out activity recognition and prediction in the homes, investigate data fusion combining binary data and depth video data, and carry out higher level activity recognition by utilizing movement and location information.

\subsection*{Acknowledgement}

The authors would like to thank the residents and the housekeepers at the seniors' care unit Sk\o yen Omsorg+; Torhild Holthe and Dr. Anne Lund (OsloMet) for recruiting participants for the trial and communicating with the residents throughout the trial; Dejan Kruni\'{c} and \O yvind Width (Sensio AS) for installations of the sensors; Oda Olsen Nedrejord and Wonho Lee (OsloMet) for contributions to the video data work; Prof. Jim T\o rresen (University of Oslo) for valuable inputs; and the rest of the participants of the Assisted Living Project for a fruitful interdisciplinary collaboration. 

%
%
%

\bibliographystyle{splncs04}
\bibliography{bin}

\begin{thebibliography}{10}
\providecommand{\url}[1]{\texttt{#1}}
\providecommand{\urlprefix}{URL }
\providecommand{\doi}[1]{https://doi.org/#1}

\bibitem{rm}
{RoomMate}. \url{https://www.roommate.no/}, [Online; accessed 16-November-2018]

\bibitem{Alam2012a}
Alam, M.R., Reaz, M.B., {Mohd Ali}, M.A.: {SPEED: An inhabitant activity
  prediction algorithm for smart homes}. IEEE Transactions on Systems, Man, and
  Cybernetics Part A:Systems and Humans  \textbf{42}(4),  985--990 (2012)

\bibitem{Bataineh2011ACS}
Bataineh, K.M., Najia, M., Saqera, M.: A comparison study between various fuzzy
  clustering algorithms (2011)

\bibitem{Blackman2016}
Blackman, S., Matlo, C., Bobrovitskiy, C., Waldoch, A., Fang, M.L., Jackson,
  P., Mihailidis, A., Nyg{\aa}rd, L., Astell, A., Sixsmith, A.: {Ambient
  Assisted Living Technologies for Aging Well: A Scoping Review}. Journal of
  Intelligent Systems  \textbf{25}(1),  55--69 (2016)

\bibitem{pelham1990}
Box, G.E.P., Jenkins, G.: Time Series Analysis, Forecasting and Control.
  Holden-Day, Inc., San Francisco, CA, USA (1990)

\bibitem{FlaviaDiasCasagrandeJimTørresen2018}
Casagrande, F.D., T{\o}rresen, J., Zouganeli, E.: {Sensor Event Prediction
  using Recurrent Neural Network in Smart Homes for Older Adults}. In:
  International Conference on Intelligent Systems (2018)

\bibitem{Casagrande2019}
Casagrande, F.D., T{\o}rresen, J., Zouganeli, E.: {Comparison of Probabilistic
  Models and Neural Networks on Prediction of Home Sensor Events}. In: Accepted
  at International Joint Conference on Neural Networks (2019)

\bibitem{Casagarnde}
Casagrande, F.D., Zouganeli, E.: {Occupancy and Daily Activity Event Modelling
  in Smart Homes for Older Adults with Mild Cognitive Impairment or Dementia}.
  In: Proceedings of The 59th Conference on Simulation and Modelling (SIMS 59).
  pp. 236--242 (2018)

\bibitem{Elhady}
Elhady, N.E., Provost, J.: {A Systematic Survey on Sensor Failure Detection and
  Fault-Tolerance in Ambient Assisted Living}. Sensors  \textbf{18}(1991), ~19
  (2018)

\bibitem{Cookc}
Gopalratnam, K., Cook, D.J.: {Online Sequential Prediction via Incremental
  Parsing: The Active LeZi Algorithm}. IEEE Intelligent Systems  \textbf{22}(1)
  (2007)

\bibitem{Hochreiter1997}
Hochreiter, S., Schmidhuber, J.: {Long Short-Term Memory}. Neural Computation
  \textbf{9}(8),  1735--1780 (1997)

\bibitem{Joshi2013ModifiedKF}
Joshi, K.D., Nalwade, P.S.: Modified k-means for better initial cluster centres
  (2013)

\bibitem{Lotfi2012}
Lotfi, A., Langensiepen, C., Mahmoud, S.M., Akhlaghinia, M.J.: {Smart homes for
  the elderly dementia sufferers: Identification and prediction of abnormal
  behaviour}. Journal of Ambient Intelligence and Humanized Computing
  \textbf{3}(3),  205--218 (2012)

\bibitem{Mahmoud2013}
Mahmoud, S., Lotfi, A., Langensiepen, C.: {Behavioural pattern identification
  and prediction in intelligent environments}. Applied Soft Computing Journal
  \textbf{13}(4),  1813--1822 (2013)

\bibitem{poisson2016}
{Mahmud}, T., {Hasan}, M., {Chakraborty}, A., {Roy-Chowdhury}, A.K.: A poisson
  process model for activity forecasting. In: 2016 IEEE International
  Conference on Image Processing (ICIP). pp. 3339--3343 (Sep 2016)

\bibitem{Minor2016}
Minor, B., Cook, D.J.: {Forecasting occurrences of activities}. Pervasive and
  Mobile Computing  (2016)

\bibitem{He2015}
Nazerfard, E., Cook, D.J.: {CRAFFT: An Activity Prediction Model based on
  Bayesian Networks}  \textbf{33}(4),  395--401 (2015)

\bibitem{reeder}
Reeder, B., Meyer, E., Lazar, A., Chaudhuri, S., Thompson, H.J., Demiris, G.:
  {Framing the evidence for health smart homes and home-based consumer health
  technologies as a public health intervention for independent aging: a
  systematic review}. International journal of medical informatics
  \textbf{7}(82),  565--579 (2013)

\bibitem{Shi2015}
Shi, X., Chen, Z., Wang, H., Yeung, D.Y., Wong, W.K., Woo, W.C.: {Convolutional
  LSTM Network: A Machine Learning Approach for Precipitation Nowcasting}
  (2015)

\bibitem{wangaction}
{Wang}, P., {Li}, W., {Gao}, Z., {Zhang}, J., {Tang}, C., {Ogunbona}, P.O.:
  Action recognition from depth maps using deep convolutional neural networks.
  IEEE Transactions on Human-Machine Systems  \textbf{46}(4),  498--509 (Aug
  2016)

\bibitem{Wu2017}
Wu, S., Rendall, J.B., Smith, M.J., Zhu, S., Xu, J., Wang, H., Yang, Q., Qin,
  P.: {Survey on Prediction Algorithms in Smart Homes}. IEEE Internet of Things
  Journal  \textbf{4}(3),  636--644 (2017)

\bibitem{Xia2012a}
Xia, L., Chen, C.C., Aggarwal, J.K.: {View invariant human action recognition
  using histograms of 3D joints}. In: IEEE Computer Society Conference on
  Computer Vision and Pattern Recognition Workshops. pp. 20--27. IEEE (jun
  2012)

\bibitem{Yang2012b}
Yang, X., Zhang, C., Tian, Y.: {Recognizing actions using depth motion
  maps-based histograms of oriented gradients}. In: Proceedings of the 20th ACM
  international conference on Multimedia - MM '12. p.~1057 (2012)

\bibitem{Zhang2017}
Zhang, S., Wei, Z., Nie, J., Huang, L., Wang, S., Li, Z.: {A Review on Human
  Activity Recognition Using Vision-Based Method}. Journal of Healthcare
  Engineering  \textbf{2017} (2017)

\bibitem{Zouganeli2017}
Zouganeli, E., Casagrande, F., Holthe, T., Lund, A., Halvorsrud, L., Karterud,
  D., Flakke-Johannessen, A., Lovett, H., M{\o}rk, S., Str{\o}m-Gundersen, J.,
  Thorstensen, E., Norvoll, R., Meulen, R., Kennedy, M.R., Owen, R., Ladikas,
  M., Forsberg, E.M.: {Responsible development of self-learning assisted living
  technology for older adults with mild cognitive impairment or dementia}.
  ICT4AWE 2017 - Proceedings of the 3rd International Conference on Information
  and Communication Technologies for Ageing Well and e-Health (Ict4awe),
  204--209 (2017)

\end{thebibliography}

\end{document}